\documentclass[letterpaper, 10 pt, conference]{ieeeconf}  
\usepackage{graphicx}
\usepackage[ruled,vlined]{algorithm2e}
\usepackage{algorithmicx}
\usepackage{amsmath}
\usepackage{mathtools}
\usepackage{booktabs}
\usepackage{array}
\usepackage{subcaption}
\usepackage{booktabs}
\usepackage{tabularx}
\usepackage{multirow}
\usepackage{threeparttable}
\usepackage{float}
\usepackage{svg}
\usepackage{amssymb}
\usepackage{amsfonts}
\usepackage{float}
\usepackage{comment}
\usepackage{url} 
\usepackage{balance}
\usepackage{cite}
\usepackage{multicol}
\usepackage[noend]{algpseudocode}
\usepackage[colorlinks=true, linkcolor=blue, citecolor=red, urlcolor=red]{hyperref}
\makeatletter
\let\NAT@parse\undefined
\makeatother

\IEEEoverridecommandlockouts                             

\title{\LARGE \bf
H-MaP: An Iterative and Hybrid Sequential Manipulation Planner
}
\author{Berk Cicek$^{1,\dagger}$, Arda Sarp Yenicesu$^{1,\dagger}$, Cankut Bora Tuncer$^{1,\dagger}$, Kutay Demiray$^{1}$, and Ozgur S. Oguz$^{1}$
\thanks{$^{\dagger}$These authors contributed equally to this work.}
\thanks{$^{1}$Dept. of Computer Engineering, Bilkent University.}
\thanks{*This work was supported by TUBITAK under 2232 program with project number 121C148 (``LiRA").}
\thanks{\textbf{Corresponding author:} Berk Cicek, Department of Computer Engineering, Bilkent University, 06800 Bilkent, Ankara, Turkey. Email: {\tt\small berk.cicek@bilkent.edu.tr}}%
\thanks{This work has been submitted to the IEEE for possible publication. Copyright may be transferred without notice, after which this version may no longer be accessible.}
}

\begin{document}

\maketitle
\thispagestyle{empty}
\pagestyle{empty}

\begin{abstract}
This paper introduces H-MaP, a hybrid sequential manipulation planner that addresses complex tasks requiring both sequential actions and dynamic contact mode switches. Our approach reduces configuration space dimensionality by decoupling object trajectory planning from manipulation planning through object-based waypoint generation, informed contact sampling, and optimization-based motion planning. This architecture enables handling of challenging scenarios involving tool use, auxiliary object manipulation, and bimanual coordination. Experimental results across seven diverse tasks demonstrate H-MaP's superior performance compared to existing methods, particularly in highly constrained environments where traditional approaches fail due to local minima or scalability issues. The planner's effectiveness is validated through both simulation and real-robot experiments.
\url{https://sites.google.com/view/h-map/}
\end{abstract}


\section{INTRODUCTION}
Sequential manipulation is fundamental to robotics, enabling robots to execute complex, multi-step processes by chaining individual actions. Sequential manipulation is fundamental to robotics, and developing robust capabilities requires addressing several significant challenges.

A primary challenge lies in creating generalized methods that can address diverse task scenarios. Consider a task where a robot must pick up a stick and then use it to push an object (Fig.~\ref{fig:intro} top-right). The solution framework must effectively handle such combinations of distinct actions while maintaining flexibility across different task configurations.

While~\cite{toussaint2018differentiable} made progress in addressing this challenge through an optimization-based approach using kinematic mode abstraction, their method faces notable limitations. In particular, the optimization framework is susceptible to becoming trapped in local minima when operating in constrained environments, compromising its robustness. This limitation becomes evident in scenarios such as pushing an object through a tunnel and retrieving it from the other side (Fig.\ref{fig:actions}-a). 

Such manipulation tasks in constrained environments necessitate dynamic contact mode switches. Recent advances in model-based dexterous manipulation have addressed this requirement through solvers that explicitly incorporate these mode transitions. These approaches typically achieve active mode switching through contact mode sampling, which helps identify global solutions~\cite{pang2023global,cheng2022contact,Lee2015HierarchicalPF}.

However, current dexterous manipulation methods face significant scalability challenges. This limitation stems from the high dimensionality of their configuration spaces, which must account for both manipulator and object states simultaneously. As a result, these methods struggle to generate extended manipulation sequences efficiently. Furthermore, their applicability to complex scenarios involving tool use and auxiliary object manipulation remains largely unexplored.

{
\setlength{\belowcaptionskip}{-18pt}
\begin{figure}[!t]
    \centering
    \includegraphics[width=0.4\textwidth]{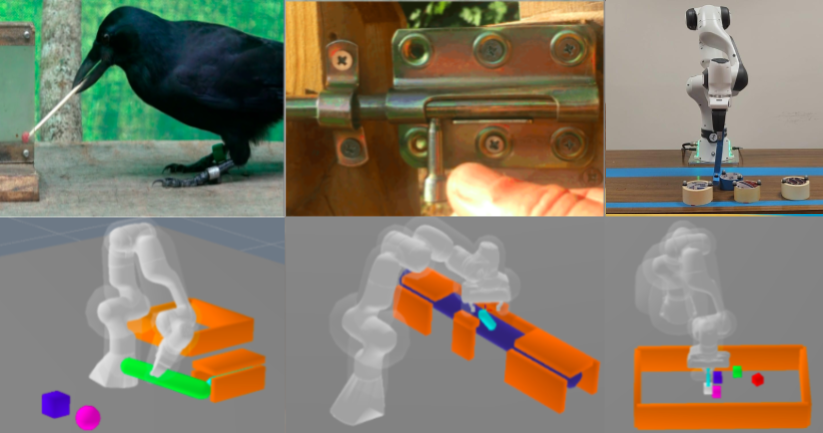}
    \captionsetup{font=footnotesize}
    \caption{(Top) Examples of tool manipulation in nature: a crow using a tool to obtain food\cite{mccoy2019new} and manipulating a sliding bolt latch\protect\footnotemark, alongside a Franka Panda robot performing an analogous obstacle removal task. (Bottom) Complex manipulation scenarios solved by our proposed approach: tool-assisted object retrieval through pushing and picking, latch mechanism manipulation, and tool-based obstacle removal.}
    \label{fig:intro}
\end{figure}
}
Therefore, developing a modular and robust manipulation planner requires addressing two fundamental criteria: the ability to effectively handle sequences of distinct actions and the capacity to generate dynamic contact mode switches.\footnotetext{\url{https://en.wikipedia.org/wiki/Latch}}

Drawing inspiration from human physical manipulation capabilities offers promising directions for addressing these sequential manipulation challenges. The human brain demonstrates remarkable capabilities: anticipating outcomes of environmental interactions~\cite{wolpert2001motor}, predicting object trajectories~\cite{kawato1999internal}, and simulating action consequences~\cite{osiurak2016tool}. Furthermore, humans naturally determine optimal contact points during object manipulation~\cite{kleinholdermann2013human}.  These biological insights helped inform the development of our approach.

This paper introduces an Iterative and Hybrid Sequential Manipulation Planner (H-MaP) that combines optimization-based methods~\cite{toussaint2014newton,Toussaint2017KOMO} with sampling-based approaches to enable complex constrained manipulation planning. The key contribution lies in our object-based trajectory planning methodology and the joint optimization of manipulator and object motion. This hybrid approach effectively addresses two critical challenges in constrained manipulation: the dimensionality explosion of configuration space and the susceptibility to local minima.

H-MaP employs an iterative planning strategy based on two fundamental concepts: \textit{waypoints} and \textit{contact points}, representing discrete object poses and physical interaction points, respectively. The iterative computation of contact points between waypoints enhances solver robustness while decomposing complex tasks into manageable segments. Our proposed informed contact sampling methodology ensures the feasibility of contact points relative to waypoints, while contact modes guide the dynamic adjustment of robot-object interactions during motion planning. This integration of waypoint and contact point information within an optimization framework enables the solver to execute sophisticated sequential manipulation tasks.

Our key contributions are summarized as follows:
\begin{itemize}
\item A hybrid manipulation planner that combines sampling- and optimization-based methods to solve complex constrained tasks, including scenarios involving auxiliary objects and tool use.
\item An iterative planning approach using waypoints and contact points that enhances both modularity and robustness in physical sequential manipulation planning.
\item A novel contact sampling methodology that enables effective integration of decoupled waypoint and contact point representations.
\end{itemize}
We evaluate our proposed planner against existing approaches across seven diverse constrained manipulation tasks, including both single-arm and bimanual scenarios, to demonstrate its scalability. The effectiveness of our method is validated through both simulation studies and real-robot experiments, confirming its practical applicability. Additionally, we provide a dataset of successful manipulation examples used for training our contact point inference model, which can serve as a benchmark for future research in learning-based manipulation planning.

\section{RELATED WORK}
\subsection{Optimization-Based Manipulation Planning}
Manipulation planning encompasses the determination of motion sequences and actions that enable robots to interact effectively with their environment. Optimization-based approaches in this domain seek to generate optimal trajectories by minimizing or maximizing objective functions subject to specific constraints, incorporating criteria such as path efficiency, energy conservation, and obstacle avoidance.

Several significant optimization frameworks have emerged in this field. CHOMP~\cite{zucker2013chomp} and its derivatives~\cite{dragan2011manipulation,dragan2017learning} utilize covariant gradient descent for cost functional optimization. STOMP~\cite{kalakrishnan2011stomp} addresses non-differentiable costs through stochastic sampling approaches, while TrajOpt~\cite{schulman2014motion} implements sequential quadratic programming with continuous-time collision checking. Logic-Geometric Programming (LGP)\cite{toussaint2015logic}, built upon KOMO\cite{toussaint2014newton}, extends these concepts by integrating logic-controlled constraints within geometric optimization problems.

Despite their computational efficiency, these optimization-based methods share a fundamental limitation: they are inherently restricted to finding locally optimal solutions, a limitation that becomes particularly problematic in constrained environments where the solution space is complex and discontinuous.
\subsection{Sampling-Based Manipulation Planning}
Sampling-based manipulation planning approaches explore high-dimensional spaces through random sampling to generate diverse action sequences, enabling complex robotic interactions. Notable methods such as CBiRRT~\cite{berenson2009manipulation}, IMACS~\cite{kingston2019exploring}, and RMR*\cite{schmitt2017optimal} focus their search on combinations of primitive actions. However, these methods' reliance on predefined motion primitives inherently constrains robotic dexterity due to their finite solution sets\cite{nakamura2017complexities}.

Recent advances, exemplified by CMGMP~\cite{cheng2022contact}, have enhanced manipulation capabilities through automatic generation of motion primitives. By integrating rapidly exploring random trees (RRT)~\cite{lavalle1998rapidly} with dynamic simulations that consider contact modes (sticking/sliding) and object dynamics, these methods successfully address challenging tasks such as flat object manipulation and sliding in constrained spaces.

While sampling-based approaches offer robust solutions, they face significant computational challenges stemming from the high dimensionality of configuration spaces and the necessity of post-processing for trajectory smoothing.
\subsection{Constrained Sequential Manipulation Planning}
Constrained Sequential Manipulation Planning requires generating action sequences for robots to accomplish complex tasks while maintaining compliance with environmental and task-specific constraints. The challenge in constrained manipulation stems from environmental limitations that restrict both object movement and available contact modes. Success in these scenarios critically depends on precise robot-object contact points, with an additional layer of complexity introduced by tasks requiring dynamic contact mode transitions.

Recent approaches have addressed different aspects of these challenges. Sampling-based methods such as CMGMP~\cite{cheng2022contact} and~\cite{pang2023global} successfully handle dynamic contact mode switches but lack comprehensive sequential manipulation capabilities. Conversely, existing sequential manipulation planners~\cite{toussaint2018differentiable, bayraktar2023solving} achieve task sequencing but cannot actively manage contact mode transitions. This paper addresses this gap by introducing a novel solver that integrates both constrained and sequential manipulation planning capabilities, representing the first comprehensive approach to unified constrained sequential manipulation.
{
\setlength{\belowcaptionskip}{-20pt}
\begin{figure*}[!th]
  \centering
  \resizebox{1\textwidth}{!}{\includegraphics[]{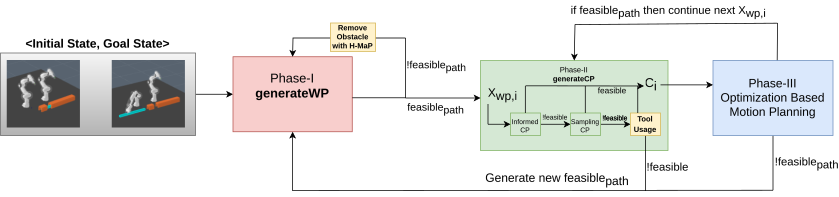}}
  \captionsetup{font=footnotesize}
  \caption{System flowchart of H-MaP's three-phase architecture: (I) Bi-RRT-based waypoint generation with obstacle handling, (II) hybrid contact point determination through learning and sampling, and (III) optimization-based motion planning. The system supports both standard end-effector ($manipulator_0$) and tool-based manipulations ($manipulator_n$), enabling dynamic replanning and recursive obstacle handling for single and multi-tool scenarios.}
  \label{fig:modif}
\end{figure*}
}
\section{PRELIMINARIES}
We first explain the k-order motion optimization formulation, KOMO \cite{toussaint2014newton}, a Non-Linear Programming (NLP) solver. Subsequently, we introduce our problem formulation.
\subsection{Optimization Based Motion Planning with KOMO}
KOMO focuses on optimizing robot trajectories through a general problem formulation that integrates constraints on movement. Therefore, it aims to find the most efficient path while complying with predefined physical and environmental constraints by solving the k-order non-linear optimization problem formulated in (\ref{eq:KOMO}).
\begin{equation}
\begin{aligned}
\label{eq:KOMO}
\min_{x_{0:T}} \quad & \sum_{t=0}^{T} f_t(x_{t-k:t})^T f_t(x_{t-k:t})\\
\text{s.t.} \quad & \forall t : g_t(x_{t-k:t}) \leq 0, \quad h_t(x_{t-k:t}) = 0.
\end{aligned}
\end{equation}

Here, $x_t \in \mathbb{R}^n$ denotes a configuration of robot and objects, while $x_{0:T} = (x_0, \ldots, x_T)$ represents a trajectory spanning a horizon $T$.
The expression $x_{t-k:t} = (x_{t-k}, \ldots, x_{t-1}, x_t)$ represents sequences of $k + 1$ consecutive states, indicating the progression of states over time within the given framework.
The functions $f_t(x_{t-k:t})$, $g_t(x_{t-k:t})$, and $h_t(x_{t-k:t})$, each mapping to $\mathbb{R}^{d_t}$, $\mathbb{R}^{m_t}$, and $\mathbb{R}^{l_t}$ respectively, are designed as arbitrary, first-order differentiable, non-linear, k-order, vector-valued functions. 
These functions serve to define cost metrics or set equality/inequality constraints for each timestep $t$.
\subsection{Problem Formulation}
We consider the problem of prehensile and non-prehensile manipulation of rigid objects and tools with a robotic manipulator.
For problem formulation, we adopt a constraint optimization framework with kinematic modes as outlined in \cite{toussaint2015logic}. We use KOMO as an NLP solver for motion optimization and incorporates kinematic switches for sequential manipulation.

We extend the problem formulation of \cite{toussaint2018differentiable,toussaint2020describing} by discretizing the path optimization.
We optimize a path ${x : [0, I] \rightarrow \mathcal{X}}$ consisting of ${I \in \mathbb{N}}$ discrete states or waypoints.
For each of the consecutive waypoints, we optimize a sub-path ${x_i : [0, K_i T] \rightarrow \mathcal{X}_i}$ consisting of ${K_i \in \mathbb{N}}$ phases or modes.
Collection of each sub-paths, ${\cup_{0}^{I} x_i}$, gives the actual full path $x$.

We define the optimization problem for each segment path \(x_i\) by employing \(s_{k,i}\), which specifies the constraints and cost parameters for a particular phase \(k_i\) along the path. 
A sequence of discrete variables \(s_{1:K_i}\) is designated as a \textit{skeleton}~\cite{toussaint2015logic}.
For each sub-path $x_i$, a different skeleton can be used.
The configuration space, denoted as \(\mathcal{X}\), spans both the \(n\)-dimensional joint space of the robot and the poses of \(m\) rigid objects within \(SE(3)^m\), starting from an initial configuration \(x_0\) within \(\mathcal{X}\). 

Given a skeleton $s_{1:K_i}$, where $\forall_{i} \in [0,I]$, we solve the optimization problem
{\small
\begin{align}
\label{eq:H-MaP}
\min_{{x_i}:[0,K_i T]\rightarrow X} & \int_{0}^{K_i T} f_{\text{path}}(\Bar{x}_i(t),s_{k,i}) \, dt + f_{\text{goal}}(x_i(T),s_{k,i})  \tag{2a} \\
\text{s.t.} \quad & x_i(0) = x_{0,i},  h_{\text{goal}}(x_i(T)) = 0, \, g_{\text{goal}}(x_i(T)) \leq 0,\tag{2b} \\
& \begin{aligned}
\forall_{t} \in [0,T]: \quad & h_{\text{path}}(\Bar{x}_i(t), s_{k,i}(t)) = 0, \\
& g_{\text{path}}(\Bar{x}_i(t), s_{k,i}(t)) \leq 0,
\end{aligned} \tag{2c} \\
& \begin{aligned}
\forall_{k} \in \{1,...,K_i\}: \quad & h_{\text{switch}}(\hat{x}_i(t_k), s_{k-1,i}, s_{k,i}) = 0, \\
& g_{\text{switch}}(\hat{x}_i(t_k), s_{k-1,i}, s_{k,i}) \leq 0,
\end{aligned} \tag{2d}
\end{align}
}Within this model, the path constraints, labeled as \((h, g)_{\text{path}}\), rely on the comprehensive state \(\Bar{x}(t) = (x(t), \dot{x}(t), \ddot{x}(t))\) to validate that the trajectory adheres to kinematic constraints, ensuring feasible motion paths.
The cost function for the path, \(f_{\text{path}}\), incorporates control efforts, specifically chosen as the sum of squared joint accelerations to evaluate and minimize the effort required for path execution.
The terms \((f, h, g)_{\text{goal}}\) denote a variety of objectives or conditions that the final state of the system is required to achieve or satisfy, ensuring the fulfillment of specified end goals.               
The terms \((h, g)_{\text{switch}}\) specify the constraints that facilitate smooth transitions between different operational modes, \(s_{k-1}\) and \(s_k\), relying on the extended state \(\hat{x} = (x, \dot{x})\) to ensure these transitions are both feasible and differentiable.
\section{ITERATIVE HYBRID MANIPULATION PLANNER}
\subsection{Overview}
Fig.~\ref{fig:modif} illustrates our iterative Hybrid sequential Manipulation Planner (H-MaP), which comprises three fundamental phases: waypoint generation, contact point sampling, and optimization-based motion planning.

The waypoint generation phase constructs a sequence of intermediate poses \( \{ X_\textrm{wp} \}^{(0:I)} \) for the target object, defining discrete states from the initial pose (\( X_0^{\text{obj}} \)) to the goal pose (\( X_G^{\text{obj}} \)). This decomposition strategy serves two key purposes: it breaks down complex manipulation tasks into manageable segments and enhances scalability by initially planning in object space, independent of robot configurations.

In the contact point sampling phase, we determine the feasible physical interaction locations between the manipulator and the target object at the intermediate poses generated by the waypoint generation phase. By default, the manipulator ($ID_{\text{manip}}$) is the robot's end-effector, except in tool-use scenarios where the tool serves as the manipulator (Fig.~\ref{fig:modif}). We enhance the contact sampling methodology from \cite{pang2023global} with a learning-informed approach to efficiently identify feasible contacts. This phase guides kinematic mode transitions and helps avoid local minima while enabling an initially decoupled object-manipulator motion planning.

Leveraging the sampled waypoints, our approach first determines feasible object trajectories in constrained environments. The final phase then transforms these trajectories into robot motion plans by merging the decoupled waypoints and contact points using an optimization-based solver~\cite{toussaint2014newton}. This architecture enables H-MaP to generate plans between consecutive waypoints (addressing sequential action challenges) while utilizing sampled contact points to manage dynamic contact mode transitions. The complete iterative algorithm is presented in Algorithm~\ref{alg:H-MaP}.

For scenarios requiring obstacle removal (as shown in the flowchart's recursive obstacle handling), H-MaP recursively plans the motion of obstacle objects to predefined removal poses. Each obstacle is treated as a new target object for a separate instance of H-MaP, with the removal poses serving as goal configurations. This recursive planning continues until the path for the original target object becomes feasible. The predefined removal poses are selected to ensure they do not interfere with subsequent manipulation tasks while remaining within the robot's workspace.
\begin{algorithm}[!ht]
\caption{H-MaP Algorithm}
\label{alg:H-MaP}
\small
\textbf{Inputs:} $X_0$ (initial configuration), $X_G^{\text{obj}}$ (goal pose), $\tau$ (goal threshold), $\text{j}_{\textrm{max}}$ (max iterations)\\
\textbf{Output:} $path$ (motion plan)\\
$feasible_{\textrm{path}} \gets \text{False}$\\
\While{$!\text{feasible}_{\textrm{path}}$}{
    \tcp*[h]{WP generation phase}\\
    $X_\textrm{wp} \gets \text{\textbf{ObjectBasedRRT}}(X_0, X_G^{\text{obj}})$\\
    \tcp*[h]{$I$: number of waypoints, len($X_\textrm{wp}$)}\\
    \tcp*[h]{Main Loop, initialized with i=0}\\
    \While{$(\|X_{i}^{\text{obj}} - X_G^{\text{obj}}\| > \tau) \; \land \; (\text{i} \not= I)$}{
        $\text{j} \gets 0$ \Comment{reset iteration number of inner loop}\\
        $feasible_{\textrm{path}} \gets \text{False}$ \Comment{re-initialize the flag}\\
        \While{$(!\text{feasible}_{\textrm{path}}) \land (\text{j} \not= \text{j}_{\textrm{max}})$}{
            \tcp*[h]{CP generation phase}\\
            \tcp*[h]{$ID_{\textrm{obj}}$: object of interest}\\
            \tcp*[h]{$ID_{\textrm{manip}}$: manipulator}\\
            $C$ $\gets$ \textbf{GenerateCP}($X_{i}, ID_{\textrm{obj}}, ID_{\textrm{manip}}$)\\
            \tcp*[h]{Optimization phase}\\
            \tcp*[h]{$X_{\textrm{wp},i}$: selected WP as sub-goal}\\
            $X_{i+1}$, path(i), $feasible_{\textrm{path}}$ $\gets \text{\textbf{KOMO}}(X_{i}$, C, $ID_{\textrm{manip}},$ $X_{\textrm{wp},i}$)$)$\\
            j $\gets$ j + 1
        }
        \If{$feasible_{\textrm{path}}$}{
            $path \gets path \cup \{\text{path}(i)\}$\\
            i $\gets$ i + 1 
        } 
        \Else{
            \textbf{break}
        }
    }   
}
\textbf{return} $path$
\end{algorithm}
\subsection{Waypoint Generation}
\label{sec:waypoint_gen}
The waypoint generation phase reduces problem complexity by focusing solely on the target object's ($ID_{\textrm{obj}}$) trajectory, thereby limiting the configuration space dimensionality. We employ Bi-RRT~\cite{bi-rrt} to compute the object's path, treating it as a freely moving body in 3D space without considering robot-object interactions. The only constraints considered during this phase are the geometric limits imposed by the task environment and objects.

The generated path undergoes optimization to ensure both optimality and smoothness while maintaining the probabilistic completeness inherited from Bi-RRT. From this optimized path, we extract waypoints $X_\textrm{wp}$ that represent the target object's trajectory. The complete waypoint generation algorithm is shown in Algorithm~\ref{alg:WPgeneration}.

\begin{algorithm}[!ht]
\caption{ObjectBasedRRT}
\label{alg:WPgeneration}
\small
\textbf{Inputs:} $X_0$ (initial configuration), $X_G^{\text{obj}}$ (goal pose)\\
\textbf{Output:} $X_\textrm{wp}$ (List of Waypoints)\\
\tcp*[h]{Find List of Waypoints}\\
\While{$path_{\textrm{sampled}} \not= Feasible$}{
    $path_{\textrm{sampled}}$ $\gets$ \textbf{RRT}($X_0^{\text{obj}}$ ,  $X_G^{\text{obj}}$)
}
$path_{\textrm{optimized}}$ $\gets$ \textbf{KOMO}($path_{\textrm{sampled}}$)\\ 
\tcp*[h]{Extract WPs from the path}\\
$X_\textrm{wp}$ $\gets$ \textbf{ExtractWPs}($path_{\textrm{optimized}}$)\\ 
\textbf{return} $X_\textrm{wp}$
\end{algorithm}
 
\subsection{Contact Point Generation}
\label{sec:CPgen}
To generate a feasible motion plan, we must establish physical interactions between the robot and the target object's waypoint trajectory. Our approach combines learning-based inference with sampling methods to efficiently generate valid contact points.

Initially, we employ an Artificial Neural Network (ANN) to propose contact points in 6-dimensional space (3D position and Euler angles) relative to target object frame based on the current configuration space. At step k, this configuration comprises the target object pose (\( X_k^{\text{obj}} \)) and remaining object poses in the environment (\( X_0^{\text{env}} \)) (obstacles, robots, tools etc.) present in the environment, which serve as features for our ANN. The model is trained on successful manipulation examples, where labeled contacts represent feasible contact points for each waypoint-defined configuration state. Each inferred contact point undergoes feasibility verification using a NLP solver in an inverse kinematics (IK) setting to ensure the end-effector can achieve the proposed pose relative to the current intermediate waypoint. If the inferred contact is feasible, it is accepted as a valid contact point, confirming that the trajectory plan based on waypoints can be transformed into a robot motion plan.

When the proposed contact point proves infeasible, we fall back to uniform sampling over the target object's point cloud, obtained using a depth camera with known object mask. 
This hybrid approach leverages informed contact point proposals to reduce generation time while maintaining robustness through sampling-based validation. The integration of a learning model enhances both computational efficiency and scalability through supervised learning, particularly beneficial for complex manipulation scenarios. The complete contact point generation process is detailed in Algorithm~\ref{alg:CPgeneration}.
{
\setlength{\intextsep}{0pt}
\begin{algorithm}[!ht]
\caption{GenerateCP}
\label{alg:CPgeneration}
\small
\textbf{Inputs:} $X_{k}$ (configuration at step k), $ID_{\textrm{obj}}$ (selected (target) object name), $ID_{\textrm{manip}}$ (selected manipulator name)\\
\textbf{Output:} $C$ (Contact Point)\\
$C \gets \emptyset$\\
$\text{feasible}_{\textrm{contact}}$ $\gets$ false\\
\tcp*[h]{CP generation phase}\\
$C_{\textrm{informed}}$ $\gets$ $\textbf{ANN}(X_{k})$\\
\tcp*[h]{Check the feasibility of contact}\\
\If{$\textbf{KOMO}(X_{k},C_{\textrm{informed}}, ID_{\textrm{manip}}$) }{
    $C \gets C_{\textrm{informed}}$\\
    $\text{feasible}_{\textrm{contact}}$ $\gets$ true\\
}
\While{$!\text{feasible}_{\textrm{contact}}$}{
    \tcp*[h]{Extract point cloud}\\
    $PointCloud$ $\gets$ \textbf{getCameraView}($X_{k}$, $ID_{\textrm{obj}}$) \\
    \tcp*[h]{Sample contact point}\\
    $C_{\textrm{sampled}}$ $\gets$ \textbf{sampleContactPoint}(PointCloud)\\
    \tcp*[h]{Check the feasibility of contact}\\
    \If{$\textbf{KOMO}(X_{k},C_{\textrm{sampled}}, ID_{\textrm{manip}}$) }{
    $C \gets C_{\textrm{sampled}}$\\
    $\text{feasible}_{\textrm{contact}}$ $\gets$ true\\
    }
}
\textbf{return} C
\end{algorithm}
}
\subsection{Optimization-Based Motion Planning}
\label{sec:optim}
The motion planning phase employs a two-stage optimization approach. The first stage verifies contact point feasibility, as described in Section~\ref{sec:CPgen}, by determining whether the manipulator can achieve the proposed contact configuration. This verification is formulated as an inverse kinematics optimization problem:
\begin{equation}
\label{eq:opt}
\begin{aligned}
\min_{x_T} \quad & f_T(x_T)^T f_T(x_T)\\
\text{s.t.} \quad & g_T(x_T) \leq 0, \quad h_T(x_T) = 0.
\end{aligned}
\tag{3}
\end{equation}
The equality constraint $h_T(x_T) = |X_{\text{pose}}^{\text{manip}}(x_T) - C_{\text{pose}}| = 0$ enforces the touch constraint, where $C_{\text{pose}}$ represents the pose of the generated contact point and $X_{\text{pose}}^{\text{manip}}(x_T)$ denotes the manipulator's pose at configuration $x_T$. The feasibility of the final configuration $x_T$ is determined by evaluating constraint violations within the optimization process.

In the second optimization phase, motion planning is performed by optimizing from the current state to a designated sub-goal to calculate $x_i$ as formulated in Eq.(\ref{eq:opt}). We follow the constraint definitions from\cite{toussaint2018differentiable}. For contact mode switches, we employ the stable constraint:
$h_{\textrm{switch}}(s_{k-1}) = |{}^{\text{obj}}X_{\text{pose}}^{\text{manip}}(s_{k-1}) - {}^{\text{obj}}X_{\text{pose}}^{\text{manip}}(s_{k})| = 0$
which enforces a constant relative transformation between manipulator and object in the object frame, effectively adding the object to the manipulator's kinematic chain. The touch constraint:
$h_T(x_T) = |X_{\text{pose}}^{\text{manip}}(x_T) - X_{\text{pose}}^{\text{obj}}(x_T)| = 0$
ensures zero distance between manipulator and object. For trajectory optimization, we add the pose equality constraint:
$h_T(x_T) = |X_{\text{pose}}^{\text{obj}}(x_T) - X_{\text{wp},i}| = 0$


After incorporating these constraints, we solve the optimization problem. This two-phased constraint optimization ensures kinematically feasible and stable grasps while following feasible paths between waypoints. The feasibility of the resulting path is evaluated based on the solver's computed constraint violations. If the path is deemed feasible, we append the generated path $x_i$ to the overall path list ($x \cup x_i$) and proceed to the next waypoint $X_{\textrm{wp},i+1}$. However, if the path is infeasible, the solver iteratively generates new contact points $C$ at the current waypoint $X_{\textrm{wp},i}$ until finding a feasible solution, for a threshold of iterations (40 in our experiments). If contact generation fails, the algorithm falls back to waypoint generation (Sec.~\ref{sec:waypoint_gen}).

The proposed method's two-phase optimization strategy first solves an inverse kinematics problem to efficiently identify feasible contact configurations before addressing the full motion planning problem. This hierarchy enhances computational efficiency by establishing optimal final configurations before proceeding with trajectory optimization toward specific sub-goals $X_{\textrm{wp},i+1}$. The combination of feasibility verification, iterative contact point generation, and waypoint fallback ensures robustness in highly constrained environments where traditional optimization-based approaches often fail due to local minima.
\subsection{Extension to Dynamic and Bimanual Scenarios}
H-MaP can be generalized to various complex scenarios. In our experiments, we demonstrate this adaptability through dynamic and bimanual manipulation tasks. In dynamic scenarios, where random moving obstacles appear along the planned path, the agent must replan its trajectory. We handle this by reinitializing H-MaP whenever a new object is detected in the scene using predefined object masks.

For bimanual scenarios, where task completion requires cooperation between multiple robots, we extend the contact point generation methodology described in Sec.~\ref{sec:CPgen}. The extended approach predicts which robot should interact at each given contact point. If the initially selected robot cannot reach the specified contact, the system iteratively attempts the action with other available robots until finding a feasible solution.
\setlength{\belowcaptionskip}{-5pt}
\begin{figure*}[!t]
    \centering
    \includegraphics[width=0.8\textwidth]{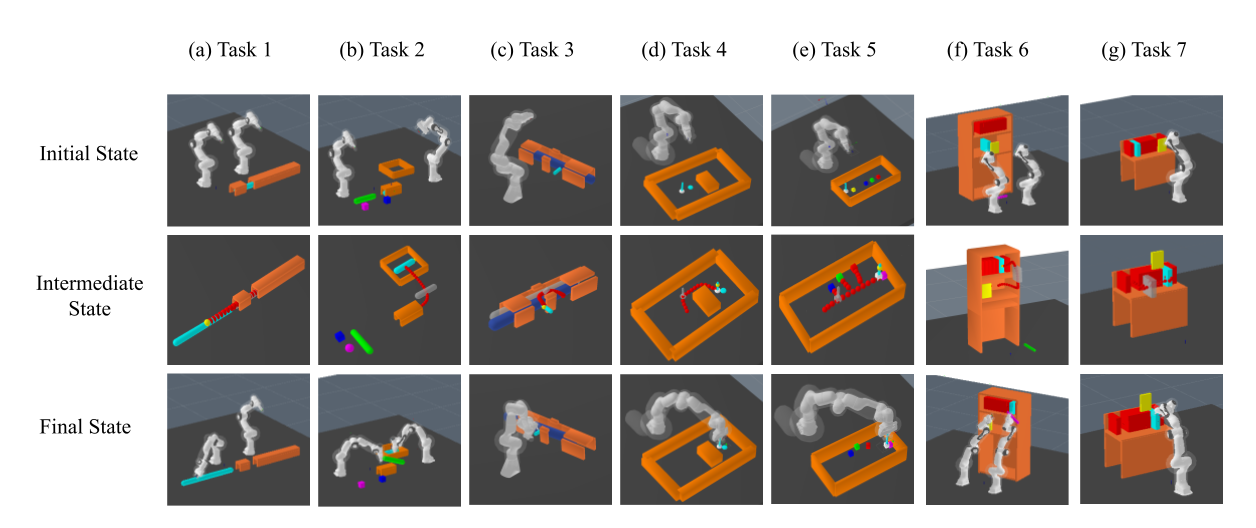}
    \captionsetup{font=footnotesize}
    \caption{From left to right: pushing an object through the tunnel; manipulating an object through the tunnel using a tool; operating a sliding latch lock; navigating an object with a tool around a fixed obstacle; and maneuvering an object with a tool while clearing movable obstacles. The top and bottom rows show initial and final configurations, respectively. The middle row displays generated waypoints (red spheres), contact points (yellow spheres), and intermediate object states (gray silhouettes) during task execution. For a complete demonstration: \url{https://sites.google.com/view/h-map/}}
    \label{fig:actions}
\end{figure*}
\section{Experiments}
\label{exp}
We evaluate H-MaP's performance on various constrained manipulation tasks, comparing it with baseline methods and detailing the dataset curation process for informed contact sampling.
\subsection{Tasks \& Scenarios}
\label{tasks}
We evaluate our algorithm across seven diverse manipulation tasks, inspired by real-world scenarios, that incorporate varying levels of complexity including multiple contact mode changes, obstacle interactions, tool use, auxiliary object manipulation, dynamic obstacles, and bimanual operations (Fig.\ref{fig:actions}). All experiments were conducted using the RAI simulator\footnote{RAI, GitHub repository, 2024, available: \url{https://github.com/MarcToussaint/rai}}, with validation on real-world implementations using Franka Panda robots. We detail each task below.

\textbf{Object through tunnel (\textit{tunnel}, Fig.~\ref{fig:actions}-a):} The robot must push an object through a tunnel and retrieve it from the opposite side in a bimanual setting. This task requires iterative pushing actions with multiple contact point adjustments to navigate the tunnel's constraints while avoiding collisions. The sequential nature of the task prevents direct object retrieval, necessitating coordinated push-then-pick manipulation between multiple robots.

\textbf{Object through tunnel with tool (\textit{tunnel with tools}, Fig.~\ref{fig:actions}-b):} A bimanual task where robots must retrieve an unreachable object from under a tunnel using an appropriate tool, followed by placing it in a box. The task complexity stems from tool selection, contact point sampling, and potential replanning due to dynamic obstacles. This extends the base tunnel task by incorporating tool-assisted manipulation.

\textbf{Sliding latch lock (\textit{bolt}, Fig.~\ref{fig:actions}-c):} The robot must manipulate a lock handle through combined vertical and horizontal movements. The task's complexity arises from non-trivial rotational movements in a highly constrained environment, challenging both waypoint generation and optimization-based planning due to local minima issues.

\textbf{Tool-assisted obstacle navigation (\textit{non-movable obstacle}, Fig.~\ref{fig:actions}-d):} The robot must use a tool to push an object to a goal position while avoiding fixed obstacles. The task requires strategic selection of multiple pushing contact points to navigate the constrained environment.

\textbf{Path clearing with tool (\textit{movable obstacles}, Fig.~\ref{fig:actions}-e):} The robot must use a tool to clear movable obstacles while pushing a target object to its goal position. The task's complexity stems from managing multiple object interactions. Initially, Bi-RRT  cannot generate feasible paths due to obstacles that require clearance.

\textbf{Bookshelf organization (\textit{bookshelf}, Fig.~\ref{fig:actions}-f):} A bimanual task requiring obstacle removal, book placement on an upper shelf, and tool-assisted alignment. The challenge lies in identifying feasible grasp points (limited to the book's middle region) and planning collision-free vertical transfers while avoiding local minima through coordinated waypoint and contact point planning.

\textbf{Single-shelf book placement (\textit{mini bookshelf}, Fig.~\ref{fig:actions}-g):} A simplified version of the bookshelf task where the robot must place a book from left to right on a single level, requiring obstacle removal and potential replanning due to dynamically appearing obstacles. The task combines precise book manipulation with adaptive planning in a constrained environment.

\textbf{Real robot validation (Fig~.\ref{fig:intro}, Fig.~\ref{fig:real-robot}):} The \textit{bookshelf}, \textit{tunnel}, and \textit{movable obstacles} tasks were replicated in a real robotic setting. To represent the state information accurately, we deployed Vicon motion capture cameras and an Intel RealSense depth camera. Plans generated in simulation were successfully executed in open-loop control using a Franka Panda robot."
{
\begin{figure}[h!]
    \centering
    \includegraphics[width=0.48\textwidth]{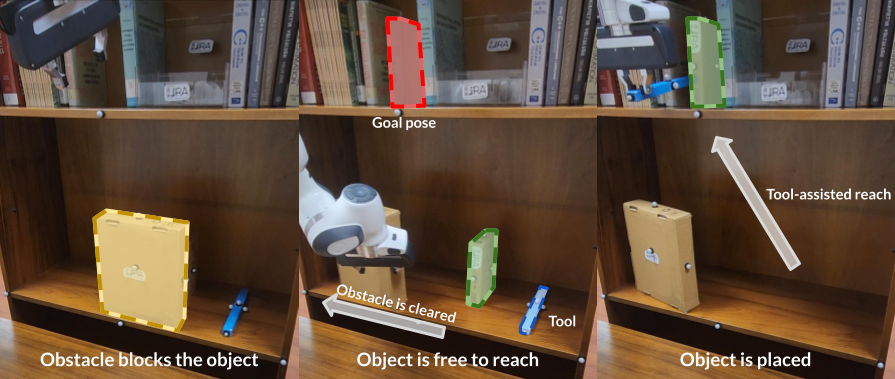}
    \captionsetup{font=small}
    \caption{\small Real robot execution of bookshelf task: obstacle removal and tool-assisted book placement.}
    \label{fig:real-robot}
    \vspace{-10pt}
\end{figure}
}

\setlength{\tabcolsep}{2pt} 

\begin{table*}[h]
\centering
\captionsetup{font=footnotesize}
\caption{Comparison of success rates (out of 10 random trials) and planning times (seconds) for successful trials. (-) indicates complete failure across all trials. (*) LGP times reflect motion planning with provided task plan skeletons, excluding task planning time, thus representing a lower bound for total LGP planning time.}
\scriptsize 
\resizebox{\textwidth}{!}{ 
\begin{tabularx}{\textwidth}{c *{14}{>{\centering\arraybackslash}X}}
\toprule
\multirow{2}{*}{Method} & \multicolumn{2}{c}{Tunnel} & \multicolumn{2}{c}{Tunnel Tool} & \multicolumn{2}{c}{Bolt} & \multicolumn{2}{c}{Puck Obs Around} & \multicolumn{2}{c}{Puck Obs Move} & \multicolumn{2}{c}{Bookshelf} & \multicolumn{2}{c}{Bookshelf Mini} \\ 
\cmidrule(lr){2-3} \cmidrule(lr){4-5} \cmidrule(lr){6-7} \cmidrule(lr){8-9} \cmidrule(lr){10-11} \cmidrule(lr){12-13} \cmidrule(lr){14-15}
 & Success & Time & Success & Time & Success & Time & Success & Time & Success & Time & Success & Time & Success & Time \\ 
\midrule
KOMO & 0/10 & - & 1/10 & 58.07$\pm$14.8 & 2/10 & 5.12$\pm$2.1 & 5/10 & 2.86$\pm$0.6 & 0/10 & - & 2/10 & 131.25$\pm$10.7 & 4/10 & 5.43$\pm$2.2 \\ 
Bi-RRT & 0/10 & - & 0/10 & - & 0/10 & - & 8/10 & 0.38$\pm$0.1 & 0/10 & - & 0/10 & - & 0/10 & - \\ 
CMGMP & 0/10 & - & 0/10 & - & 0/10 & - & 5/10 & 2.29$\pm$0.1 & 4/10 & 2.30$\pm$0.2 & 0/10 & - & 0/10 & -\\
LGP* & 5/10 & 6.13$\pm$1.9 & 4/10 & 4.19$\pm$0.6 & 2/10 & 6.24$\pm$1.2 & 10/10 & 2.48$\pm$0.7 & 8/10 & 10.43$\pm$2.2 & 3/10 & 35.3$\pm$6.0 & 9/10 & 45.86$\pm$10.6 \\ 
HMAP & 10/10 & 5.08$\pm$2.4 & 10/10 & 6.44$\pm$4.6 & 10/10 & 4.05$\pm$0.8 & 10/10 & 21.59$\pm$6.1 & 10/10 & 26.4$\pm$8.2 & 10/10 & 18.92$\pm$12.4 & 10/10 & 6.06$\pm$0.8 \\ 
\bottomrule
\end{tabularx}
}
\label{tab:comparison}
\end{table*}
\subsection{Baselines}
\label{baselines}
We compare H-MaP against KOMO~\cite{toussaint2014newton}, Bi-RRT~\cite{bi-rrt}, CMGMP~\cite{cheng2022contact}, and LGP~\cite{toussaint2015logic}. KOMO and RRT are selected as they represent the foundational optimization-based and sampling-based methods that inform our work. CMGMP is chosen to benchmark against state-of-the-art sampling methods capable of dynamic contact mode switches. LGP represents current capabilities in sequential planning. For a fair comparison, we implemented tool manipulation capabilities across all baseline methods and excluded dynamic obstacle scenarios, as they were beyond the scope of the baseline approaches. For CMGMP, we modified the point manipulator parameters to match our constrained environment requirements. In LGP's case, we provided planning skeletons containing symbolic and geometric intermediate steps, effectively making it an informed version of KOMO.
\subsection{Dataset Curation}
\label{dataset}
To train our informed contact sampling method, we curated a dataset of successful manipulation examples in task-specific simulation environments detailed in Section \ref{tasks}. HMaP contact sampling was adjusted for uniform sampling across object points, and feasibility checks were applied (Section \ref{sec:CPgen}). Generated paths were tested in simulation to ensure successful, feasible motion plans. Negative samples were excluded, leaving this for future work to enable more advanced architectures.
\subsection{Results}
\label{res}
\subsubsection{Comparative Results}
Table \ref{tab:comparison} presents the performance comparison between H-MaP and baselines across seven diverse tasks. H-MaP successfully solved all tasks across 10 random trials, while baseline methods showed various limitations.

KOMO and Bi-RRT demonstrated expected limitations: optimization-based KOMO failed in complex manipulation scenarios where manifold discontinuities disrupted local Euclidean properties, leading to local minima, while Bi-RRT failed to generate feasible paths within reasonable time limits (180 seconds). CMGMP, despite being guided by contacts, faced similar scalability challenges as Bi-RRT due to the need to incorporate full manipulator configurations throughout trajectory planning.

LGP achieved moderate success by leveraging predefined symbolic and geometric intermediate states, which effectively decomposed complex tasks into subproblems. However, its reliance on optimization-based motion planning led to failures in scenarios like the bolt task, where, similar to KOMO, it struggled due to the lack of fine-grained action definitions needed to find solutions in complex non-convex manifolds.

H-MaP demonstrates superior performance through its decoupled waypoint and contact point generation strategy, effectively integrating sampling-based and optimization-based methodologies. The learning-informed contact point generation mitigates hybrid planning overhead, often outperforming conventional approaches in planning time. While simpler tasks may benefit from single-approach methods, H-MaP achieves its primary objective of robust sequential manipulation with dynamic contact mode switches. The planner maintains efficacy in dynamic environments, successfully replanning trajectories when obstacles appear within a 0.3-unit radius (approximately 6 RRT extensions) of the manipulation path.
\subsubsection{Effectiveness of Informed Sampling with Learning}
{
\setlength{\belowcaptionskip}{-10pt}
\begin{figure}[h]
    \centering
    \includegraphics[width=0.5\textwidth]{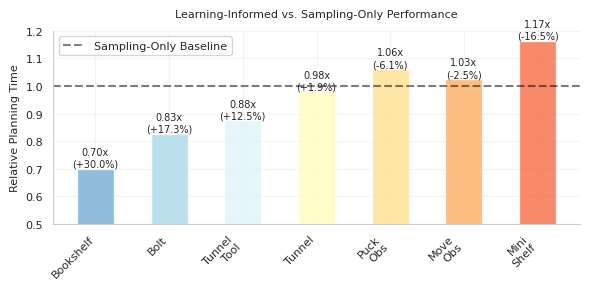}
    \captionsetup{font=footnotesize}
    \caption{Performance comparison between learning-informed and sampling-only approaches, normalized to the sampling-only baseline (1.0x). Lower bars indicate better performance, with improvements shown as percentages. Tasks are ordered by relative improvement, highlighting the effectiveness of the learning-informed approach across manipulation scenarios.}
    \label{fig:time}
\end{figure}
}
As shown in Fig. \ref{fig:time}, the learning-informed approach demonstrates a clear pattern of improvement across different manipulation tasks when compared to the sampling-only baseline. Most significantly, it reduced planning time by 30.0\% for the bookshelf task (0.70x relative time) and 17.3\% for the bolt task (0.83x relative time), which represent the most geometrically complex scenarios requiring precise contact selection. The tunnel tool task showed a 12.5\% improvement (0.88x relative time), while maintaining comparable performance in simpler scenarios such as bookshelf mini and puck obstacle tasks where single contact points suffice. This pattern suggests that our learning-informed sampling strategy provides greater benefits as task complexity increases, particularly in scenarios requiring multiple contact points and precise manipulation. The method's ability to maintain or improve performance across all task categories, without significant degradation in any scenario, demonstrates its robustness and practical utility for complex manipulation planning.
\section{DISCUSSION \& LIMITATIONS}
Our approach differs from traditional methods by focusing on sampling-based planners exclusively on object path planning rather than considering combined robot-object kinematic states. This significantly reduces configuration space dimensionality, enabling effective sequential manipulation planning. Unlike existing methods that typically handle single robot-object interactions, our approach successfully manages tool use and auxiliary object manipulation.

The results demonstrate that enhancing low-level motion planners for sequential manipulation eliminates the need for predefined motion primitives. This reduces reliance on explicit high-level action definitions, allowing implicit execution of complex actions. For instance, in the \textit{Bolt} task, our method eliminates the need to specify discrete actions like \texttt{lift up}, \texttt{pull left}, and \texttt{pull down}. However, our solver has several limitations:

\textbf{Action Limitations:} Our decoupled object and robot planning introduces certain action limitations. For example, in the flip-card scenario from \cite{cheng2022contact}, combined planners achieve a better grasp by constraining the path to the manipulator. In contrast, our approach may lead to premature pick actions, resulting in infeasible motion. Addressing this could involve informing waypoint generation with configuration-based action constraints. \textbf{Waypoint Quality:} The solver assumes RRT-generated waypoints are viable. Poor waypoint generation may lead to solver failure despite proximity-based searching. We mitigate this by regenerating waypoints when no feasible path is found. \textbf{Kinematic Focus:} The current implementation considers only kinematic systems, simplifying optimization but neglecting system dynamics. Future work could incorporate dynamics in planning and implement controllers for execution, enhancing real-world applicability.
\section{CONCLUSION}
This paper presented H-MaP, a novel hybrid sequential manipulation planner that effectively addresses two fundamental challenges in robotic manipulation: handling sequences of distinct actions and generating dynamic contact mode switches. By decoupling object trajectory planning from manipulation planning through waypoint generation and integrating learning-informed contact sampling, H-MaP significantly reduces the configuration space dimensionality while maintaining solution completeness.

Our experimental results across seven diverse manipulation tasks demonstrate H-MaP's capability to solve complex scenarios involving tool use, auxiliary object manipulation, and bimanual coordination. The planner's success in handling highly constrained environments, where traditional optimization-based methods often fail due to local minima, validates our hybrid approach. Successful real-robot implementation confirms practical applicability, while the modular architecture enables extension to dynamic and multi-robot scenarios, with future work focusing on system dynamics and backtracking strategies.




\bibliographystyle{IEEEtran}
\bibliography{cit}

\end{document}